\def\year{2022}\relax
\documentclass[letterpaper]{article} 
\usepackage{aaai22}  
\usepackage{times}  
\usepackage{helvet} 
\usepackage{courier}  
\usepackage[hyphens]{url}  
\usepackage{graphicx} 
\usepackage{natbib}  
\usepackage{caption} 
\usepackage{etoolbox,siunitx}
\usepackage{booktabs}
\usepackage{pifont}%
\usepackage[noend]{algpseudocode}
\usepackage{algorithm, algorithmicx}
\usepackage{mathtools, nccmath}
\usepackage{amsfonts}
\usepackage{amssymb}
\usepackage{mathtools} 

\usepackage[rgb,dvipsnames]{xcolor}

\DeclareSymbolFont{matha}{OML}{txmi}{m}{it}
\DeclareMathSymbol{\curledv}{\mathord}{matha}{35}

\usepackage{amsmath}
\usepackage{color}
\usepackage{soul}
\usepackage{xspace}
\usepackage[capitalize]{cleveref}
\usepackage{dsfont}
\usepackage{bm}
\usepackage{bbm}
\usepackage{nicefrac}  
\usepackage{amsthm}
\usepackage{amsfonts}
\usepackage{amssymb}
\usepackage{array}
\usepackage{mathtools}
\usepackage{caption}
\usepackage{subcaption}
\usepackage{fontawesome}
\usepackage{thmtools}
\usepackage{thm-restate}
\usepackage{lipsum}
\usepackage{pifont}
\usepackage{makecell}
\usepackage{multirow}
\usepackage{afterpage}
\usepackage{tabularx} 
\usepackage{url}
\usepackage{enumitem}  
\usepackage{wrapfig}  
\usepackage{textcomp}

\Crefname{figure}{Figure}{Figures}
\crefname{figure}{Figure}{Figures}
\crefname{table}{Table}{Tables}
\crefformat{table}{Table~#1}
\crefformat{equation}{Equation (#1)}
\Crefformat{equation}{Equation (#1)}
\crefformat{algorithm}{Algorithm~#1}
\crefformat{appendix}{Appendix~#1}
\Crefformat{appendix}{Appendix~#1}
\crefformat{appendices}{Appendices~#1}
\Crefformat{appendices}{Appendices~#1}

\newcommand\sdots{\hbox to 1em{.\hss.\hss.}} 
\DeclareMathAlphabet\mathbfcal{OMS}{cmsy}{b}{n}  
\newcommand{\smallsum}{\textstyle\sum\limits}

\newtheorem{question}{Question}

\newif\ifcomments
\commentsfalse 

\ifcomments
	\newcommand{\dXX}[1]{\color{red}DK: (#1)\color{black}\xspace}  
	\newcommand{\maXX}[1]{\color{OliveGreen}MA: (#1)\color{black}\xspace}  
	\newcommand{\XX}[1]{\color{orange}JH: (#1)\color{black}\xspace}  
	\newcommand{\mXX}[1]{\color{blue}ML: (#1)\color{black}\xspace}  
	\newcommand{\gtXX}[1]{\color{purple}GT: (#1)\color{black}\xspace}  
	\newcommand{\mrXX}[1]{\color{blue}MR: (#1)\color{black}}  
\else
    \newcommand{\dXX}[1]{}  
	\newcommand{\maXX}[1]{}  
	\newcommand{\XX}[1]{}  
	\newcommand{\mXX}[1]{}  
	\newcommand{\gtXX}[1]{}  
	\newcommand{\mrXX}[1]{}  
\fi

\title{Context-Specific Representation Abstraction for Deep Option Learning}

\author {
    Marwa Abdulhai\textsuperscript{\rm 1,2},
    Dong-Ki Kim\textsuperscript{\rm 1,2},
    Matthew Riemer\textsuperscript{\rm 2,3},
    Miao Liu\textsuperscript{\rm 2,3},\\
    Gerald Tesauro\textsuperscript{\rm 2,3},
    Jonathan P. How\textsuperscript{\rm 1,2}
}
\affiliations {
    \textsuperscript{\rm 1} MIT LIDS  \\
    \textsuperscript{\rm 2} MIT-IBM Watson AI Lab \\
    \textsuperscript{\rm 3} IBM Research\\
     abdulhai@mit.edu, dkkim93@mit.edu, mdriemer@us.ibm.com, miao.liu1@ibm.com, \\ gtesauro@us.ibm.com, jhow@mit.edu 
}

\begin{document}

\maketitle

\begin{abstract}
Hierarchical reinforcement learning has focused on discovering temporally extended actions, such as options, that can provide benefits in problems requiring extensive exploration. One promising approach that learns these options end-to-end is the option-critic (OC) framework. We examine and show in this paper that OC does not decompose a problem into simpler sub-problems, but instead increases the size of the search over policy space with each option considering the entire state space during learning. This issue can result in practical limitations of this method, including sample inefficient learning. To address this problem, we introduce Context-Specific Representation Abstraction for Deep Option Learning (CRADOL), a new framework that considers both temporal abstraction and context-specific representation abstraction to effectively reduce the size of the search over policy space. Specifically, our method learns a factored belief state representation that enables each option to learn a policy over only a subsection of the state space. We test our method against hierarchical, non-hierarchical, and modular recurrent neural network baselines, demonstrating significant sample efficiency improvements in challenging partially observable environments.
\end{abstract}

\section{Introduction} \label{sec:intro}
Hierarchical reinforcement learning (HRL) provides a principled framework for decomposing problems into natural hierarchical structures~\cite{BerliacHierachialRL2019}. 
By factoring a complex task into simpler sub-tasks, HRL can offer benefits over non-HRL approaches in solving challenging large-scale problems efficiently.
Much of HRL research has focused on discovering temporally extended high-level actions, such as options, that can be used to improve learning and planning efficiency~\cite{sutton1991, doina_thesis}. 
Notably, the option-critic (OC) framework has become popular because it can learn deep hierarchies of both option policies and termination conditions from scratch without domain-specific knowledge~\citep{bacon2016optioncritic,riemer18options}. However, while OC provides a theoretically grounded framework for HRL, it is known to exhibit common failure cases in practice including lack of option diversity, short option duration, and large sample complexity~\cite{diversity}. 

\begin{figure*}[t]
	\centering
	\includegraphics[width=0.86\linewidth]{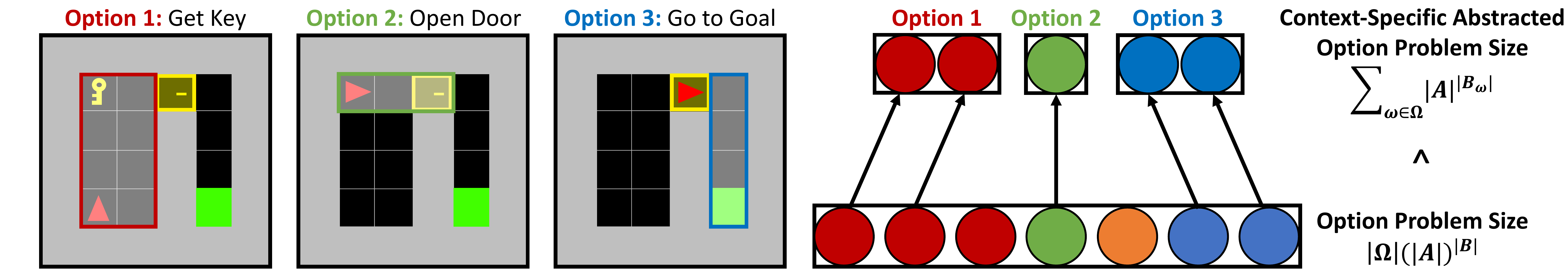}
	\caption{Temporal abstraction in this maze consists of 3 options; getting the key, opening the door, and going to the green goal. With context-specific belief state abstraction, the agent should only focus on different factors of the belief state pertaining to the highlighted sub-task. Equipped with both temporal and context-specific abstraction, our method achieves a significant reduction in the learning problem size compared to considering the entire belief state space for each option as highlighted in Remark 1.}
	\label{fig:motivation}
\end{figure*}

Our key insight in this work is that OC suffers from these problems at least partly because it considers the entire state space during option learning and thus fails to reduce problem complexity as intended. 
For example, consider the partially observable maze domain in \cref{fig:motivation}. 
The objective in this domain is to pick up the key and unlock the door to reach the green goal. 
A non-HRL learning problem in this setting can be viewed as a search over the space of deterministic policies: $|\mathcal{A}|^{|\mathcal{B}|}$, where $|\mathcal{A}|$ and $|\mathcal{B}|$ denote the size of the action space and belief state space, respectively. 
Frameworks such as OC consider the entire state space or belief state when learning options, which can be viewed themselves as policies over the entire state space, naively resulting in an increased learning problem size of $|\Omega|(|\mathcal{A}|^{|\mathcal{B}|})$, where $|\Omega|$ denotes the number of options. Note that this has not yet considered the extra needed complexity for learning policies to select over and switch options. 
Unfortunately, because $|\Omega|\!\geq\!1$, learning options in this way can then only increase the effective complexity of the learning problem.

To address the problem complexity issue, we propose to consider both temporal abstraction, in the form of high-level skills, and context-specific abstraction over factored belief states, in the form of latent representations, to learn options that can fully benefit from hierarchical learning.
Considering the maze example again, the agent does not need to consider the states going to the goal when trying to get the key (see Option 1 in \Cref{fig:motivation}). 
Similarly, the agent only needs to consider states that are relevant to solving a specific sub-task.
Hence, this context-specific abstraction leads to the desired reduction in problem complexity, and an agent can mutually benefit from having both context-specific and temporal abstraction for hierarchical learning. 

\noindent\paragraph{Remark 1} 
\textit{With context-specific abstraction for each option, the learning problem size can be reduced such that:}
\begin{align}\label{lemma1}
\begin{split}
    \sum_{\omega\in\Omega} |\mathcal{A}|^{|\mathcal{B}_{\omega}|} < |\mathcal{A}|^{|\mathcal{B}|} < |\Omega|(|\mathcal{A}|)^{|\mathcal{B}|},
\end{split}
\end{align}
where $\mathcal{B}_{\omega}\!\subset\!\mathcal{B}$ denotes the subset of the belief state space for each option (or sub-policy) $\omega$ and $|\mathcal{B}_{\omega}|\!<\!|\mathcal{B}|$. 
We further illustrate this idea in the tree diagram of \Cref{fig:motivation}, where each option maps to only a subset of the state space. 

\paragraph{Contribution.} With this insight, we introduce a new option learning framework: Context-Specific Representation Abstraction for Deep Option Learning (CRADOL). 

CRADOL provides options with the flexibility to select over dynamically learned abstracted components. This allows effective decomposition of problem size when scaling up to large state and action spaces, leading to faster learning. 
Our experiments demonstrate how CRADOL can improve performance over key baselines in partially observable settings.

\section{Problem Setting and Notation} \label{sec:background}
\paragraph{Partially Observable Markov Decision Process.}
An agent's interaction in the environment can be represented by a partially observable Markov decision process (POMDP), formally defined as a tuple $\langle \mathcal{S}, \mathcal{A}, P, \mathcal{R}, \mathcal{X}, \mathcal{O}, \gamma \rangle$~\cite{POMDPs}. 
$\mathcal{S}$ is the state space, 
$\mathcal{A}$ is the action space, 
$P$ is the state transition probability function,
$\mathcal{R}$ is the reward function, 
$\mathcal{X}$ is the observation space, 
$\mathcal{O}$ is the observation probability function, and
$\gamma\!\in\![0,1)$ is the discount factor.
An agent executes an action $a$ according to its stochastic policy $a\!\sim\!\pi_\theta(a|b)$ parameterized by $\theta\!\in\!\Theta$, where $b\!\in\!\mathcal{B}$ denotes the belief state and the agent uses the observation history up to the current timestep $x_{0:t}$ to form its belief $b_{t}$. 
Each observation $x\!\in\!\mathcal{X}$ is generated according to $x\!\sim\!\mathcal{O}(s)$.
An action $a$ yields a transition from the current state $s\!\in\!\mathcal{S}$ to the next state $s'\!\in\!\mathcal{S}$ with probability $P(s'|s,a)$, and an agent obtains a reward $r\!\sim\!\mathcal{R}(s,a)$.
An agent's goal is to find the deterministic optimal policy $\pi^*$ that maximizes its expected discounted return $\mathbb{E}_{\pi_\theta}\left[\sum_{t=0}^\infty \gamma^t r_t\right] \;\forall \theta \in \Theta$. 

\noindent\paragraph{Option-Critic Framework.} 
A Markovian option $\omega\!\in\!\Omega$ consists of a triple $\langle I_{\omega}, \pi_{\omega}, \beta_{\omega}\rangle$. 
$I_{\omega}\!\in\!\mathcal{S}$ is the initiation set, $\pi_{\omega}$ is the intra-option policy, and $\beta_{\omega}\!: \mathcal{S}\!\rightarrow\![0,1]$ is the option termination function~\cite{sutton1991}. 
Similar to option discovery methods~\cite{Mankowitz2016AdaptiveSA, determine_options_daniel, NIPS2016_c4492cbe}, we also assume that all options are available in each state.
MDPs with options become semi-MDPs~\cite{smdp} with an associated optimal value function over options $V^*_\Omega(s)$ and option-value function $Q^*_\Omega(s, \omega)$.
\citet{bacon2016optioncritic} introduces a gradient-based framework for learning sub-policies represented by an option. An option $\omega$ is selected according to a policy over options $\pi_\Omega(\omega|s)$, and $\pi_{\omega}(a|s)$ selects a primitive action $a$ until the termination of this option (as determined by $\beta_{\omega}(s)$), which triggers a repetition of this procedure. 
The framework also models another auxiliary value function $Q_U(s,\omega,a)$ for computing the gradient of $\pi_{\omega}$.

\paragraph{Factored POMDP.} 
In this work, we further assume that the environment structure can be modeled as a factored MDP in which the state space $\cal S$ is factored into $n$ variables: $\mathcal{S}\!=\!\{\mathcal{S}^1,\ldots,\mathcal{S}^n \}$, where each $\mathcal{S}^i$ takes on values in a finite domain and $i\!\in\!\{1,\ldots,n\}$~\cite{Guestrin_2003}. 
Under this assumption, a state $s\!\in\!\mathcal{S}$ is an assignment of values to the set of state variables: $s\!=\!\{s^1,\ldots,s^n\}$ such that $\mathcal{S}\!\subseteq\!\mathcal{S}^1 \times \ldots \times \mathcal{S}^n$. 
These variables are also assumed to be largely independent of each other with a small number of causal parents $\mathcal{P}^{\mathcal{S}^i}$ to consider at each timestep, resulting in the following simplification of the transition function: 
\begin{equation}\label{eq:factored_parents}
    P(s'|s,a) \approx \prod_{i=1}^n P^i(s'^i|\mathcal{P}^{\mathcal{S}^i}(s),a).
\end{equation}
This implies that in POMDP settings we can match the factored structure of the state space by representing our belief state in a factored form as well with $\mathcal{B}\!=\!\{\mathcal{B}^1,\ldots,\mathcal{B}^n \}$ and $b\!=\!\{b^1,\ldots,b^n\}$.
Another consequence of the factored MDP assumption is context-specific independence.

\paragraph{Context-Specific Independence.}
At any given timestep, only a small subset of the factored state space may be necessary for an agent to be aware of. 
As per \citet{context_definition}, we define this subset to be a \textit{context}. 
Formally, a context $Z$ is a pair $(Z,\mathcal{Z})$, where $Z\!\subseteq\!\mathcal{S}$ is a subset of state space variables and $\mathcal{Z}$ is the space of possible joint assignments of state space variables in the subset. A state $s$ is in the context $(Z,\mathcal{Z})$ when its joint assignment of variables in $Z$ is present in $\mathcal{Z}$. Two variables $Y,Y' \subseteq S \setminus Z$ are contextually independent under $(Z,\mathcal{Z})$ if $Pr(Y|Y',Z=z)=Pr(Y|Z=z) \;\forall z \in \mathcal{Z}$. This independence relation is referred to as context-specific independence (CSI) \citep{chitnis2020camps}. CSI in the state space also implies that this same structure can be leveraged in the belief state for POMDP problems. 

\section{Context-Specific Belief State Abstractions\\for Option-Critic Learning} 
We now outline how factored state space structure and context specific independence can be utilized by OC to decrease the size of the search problem over policy space. 
We first note that the size of the search problem over the policy space of OC $|\pi_{\textrm{OC}}|$ can be decomposed into the sum of the search problem sizes for each sub-policy within the architecture: 
\begin{equation}
|\pi_{\textrm{OC}}| = |\pi_\Omega| + \sum_{\omega \in \Omega} |\pi_\omega| + |\beta_\omega|.
\end{equation}
OC can provide a decrease in the size of the learning problem over the flat RL method when $|\pi_{\textrm{OC}}|\!<\!|\mathcal{A}|^{|\mathcal{B}|}$. 
This decrease is possible if an agent leverages context-specific independence relationships with respect to each option $\omega$ such that only a subset of the belief state variables are considered $\mathcal{B}_\omega\!\subset\!\mathcal{B}$, where $\mathcal{B}_\omega\!=\!\{\mathcal{B}^1,\ldots,\mathcal{B}^k \}$ and $k\!<\!n$, implying that $|\mathcal{B}_\omega|\!\ll\!|\mathcal{B}|$ because the belief state space gets exponentially smaller with each removed variable. 
The abstracted belief state $b_\omega\!\in\!\mathcal{B}_\omega$ is then sent as an input to the intra-option policy $\pi_\omega(a|b_\omega)$ and termination policy $\beta_\omega(b_\omega)$, dictating the size of the learning problem for each:
\begin{equation}
|\pi_\omega| = |\mathcal{A}|^{|\mathcal{B}_\omega|}, \qquad |\beta_\omega| = 2^{|\mathcal{B}_\omega|}.
\end{equation}
However, a challenge becomes that we must also consider the size of the learning problem for the policy over options $|\pi_\Omega|$. 
$\pi_\Omega$ cannot simply focus on an abstracted belief state in the context of an individual option $\omega$ because it must reason about the impact of all options. 
Naively sending the full belief state $b$ to $\pi_\Omega$ is unlikely to make the learning problem smaller for a large class of problems because we cannot achieve gains over the flat policy in this case if $|\Omega|$ is of comparable or greater size than $|\mathcal{A}|$. 
In this work, we address this issue by sending the policy over options its own compact abstracted belief state $b_\Omega \in \mathcal{B}_\Omega$, where $|\mathcal{B}_\Omega| < |\mathcal{B}|$. As a result, the learning problem size of $\pi_\Omega(\omega|b_\Omega)$ is $|\Omega|^{|\mathcal{B}_\Omega|}$ and the total problem size for $|\pi_{\textrm{OC}}|$ is:
\begin{equation}
|\pi_{\textrm{OC}}| = |\Omega|^{|\mathcal{B}_\Omega|} + \sum_{\omega \in \Omega} |\mathcal{A}|^{|\mathcal{B}_\omega|} + 2^{|\mathcal{B}_\omega|}.
\end{equation}
Note that $b_\Omega$ represents a different kind of belief state abstraction than $b_\omega$ in that it must consider all factors of the state space. 
$b_\Omega$, however, should also consider much less detail than is contained in $b$ for each factor, because $\pi_\Omega$ is only concerned with the higher level semi-MDP problem that operates at a much slower time scale when options last for significant lengths before terminating and does not need to reason about the intricate details needed to decide on individual actions at each timestep. 
As a result, we consider settings where $|\mathcal{B}_\Omega| \ll |\mathcal{B}|$ and $|\mathcal{B}_\omega| \ll |\mathcal{B}|$, ensuring that an agent leveraging our CRADOL framework solves a smaller learning problem than either flat RL or OC as traditionally applied.
Finally, the auxiliary value function $Q_U$, which is used to compute the gradient of $\pi_\omega$, must consider rewards accumulated across options similarly to $\pi_\Omega$. Thus, we also send its own compact abstracted belief state $b_U \in \mathcal{B}_U$ as an input to $Q_U$ such that the resulting problem size is small $|\mathcal{B}_U| \ll |\mathcal{B}|$.

\begin{figure*}[t]
	\centering
	\includegraphics[width=1.0\linewidth]{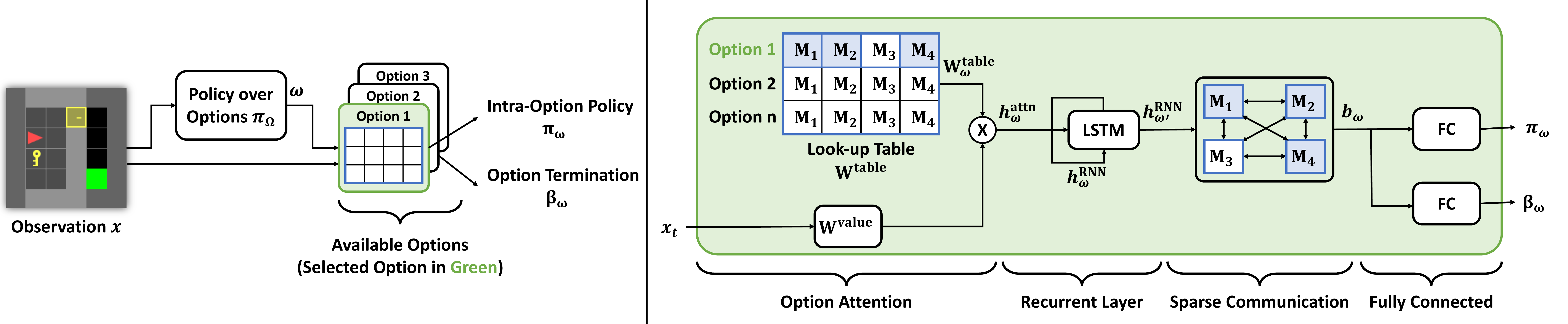}
	 \qquad
	\caption{Architecture for intra-option policy of Context-Specific Abstracted Option-Critic (CRADOL). The left side describes the overview of CRADOL, and the right side details the option learning process.}
	\label{fig:approach}
\end{figure*}

\section{Learning Context-Specific Abstractions\\with Recurrent Independent Mechanisms} 
In this section, we present a new framework, CRADOL, based on our theoretical motivation outlined for applying abstractions over belief state representations in the last section. In particular, our approach leverages recurrent independent mechanisms \citep{RIMs} to model factored context-specific belief state abstractions for deep option learning. 

\subsection{Algorithm Overview}

\cref{fig:approach} shows a high-level overview of CRADOL. CRADOL uses small LSTM networks to provide compact abstract representations for $b_\Omega$ and $b_U$. It additionally leverages a set of mechanisms to model the factored belief state $b$ while only sending a subset of this representation $b_\omega$ to each option during its operation. 

\subsection{Option Learning}
Our framework for option learning broadly consists of four modules: option attention, a recurrent layer, sparse communication, and fully connected (see \cref{fig:approach}). 
We choose to model each option based on a group of $K$ mechanisms because these mechanisms can learn to specialize into sub-components following the general structure assumed in the environment for factored POMDPs. 
This in turn allows our learned options to self-organize into groupings of belief state components.

\paragraph{Option Attention Module.}
The option attention module chooses which mechanisms out of $K$ mechanisms ($M_{1},...,M_K$) are activated for $\omega\!\in\!\Omega$ by passing the option through a look-up table $W^{\text{table}}\!\in\!\mathbb{R}^{|\Omega|\times K}$. 
The lookup table ensures a fixed context selection over time (i.e. a fixed mapping to a subset of the belief state space) by an option as considering many different subsets with the same option would be effectively equivalent to considering the entire belief state space. This would lead to a large search for policy space as previously done by OC. 
To yield the output of the attention module $h_{\omega}^{\text{attn}}$, we apply the following:
\begin{equation}\label{eq:first_layer}
h_{\omega}^{\text{attn}} = \text{softmax}\big(W^{\text{table}}_{\omega}\big)^{T}xW^{\text{value}},
\end{equation}
where $W^{\text{table}}_{\omega}\!\in\!\mathbb{R}^{1\times K}$ is the attention values for $\omega$ and $W^{\text{value}}\!\in\!\mathbb{R}^{|\mathcal{X}|\times v}$ denotes the value weight with the value size $v\in\mathbb{N}$. 

We note that a top-$k$ operation is then performed such that only $k$ mechanisms components are selected from the available $K$ mechanisms (not selected mechanisms are zero masked), which enables an option to operate under an exponentially reduced belief state space by operating over only a subset of mechanisms. 

\paragraph{Recurrent Layer Module.}
The recurrent layer module updates the hidden representation of all active mechanisms. 
Here, we have a separate RNN for each of the $K$ mechanisms with all RNNs shared across all options. 
Each recurrent operation is done for each active mechanism separately by taking input $h_{\omega}^{\text{attn}}\in\mathbb{R}^{K\times v}$ and the previous RNN hidden state $h_{\omega}^{\text{RNN}}$, in which we obtain the output of recurrent layer module $h_{\omega'}^{\text{RNN}}\!\in\!\mathbb{R}^{K\times h}$ for all mechanisms, where $h\in\mathbb{N}$ denotes RNN's hidden size. 

\paragraph{Sparse Communication Module.}
The sparse communication module enables mechanisms to share information with one another, facilitating coordination in learning amongst the options. Specifically, all active mechanisms can read hidden representations from both active and in-active mechanisms, with only active mechanisms updating their hidden states. 
This module outputs the context-specific belief state $b_{\omega}\in\mathbb{R}^{K\times h}$ given the input $h_{\omega'}^{\text{RNN}}$:

\begin{align}\label{eq:third_layer}
\begin{split}
b_{\omega}&=\big[\text{softmax}\big(\frac{W^{Q_{\text{comm}}}h_{\omega'}^{\text{RNN}}(W^{K_{\text{comm}}}h_{\omega'}^{\text{RNN}})^{T}}{\sqrt{d^{K_{\text{comm}}}}}\big)\\
&\quad\quad W^{V_{\text{comm}}}h_{\omega'}^{\text{RNN}}\big]+h_{\omega'}^{\text{RNN}},
\end{split}
\end{align}
where $W^{K_{\text{comm}}}, W^{Q_{\text{comm}}}, W^{V_{\text{comm}}}$ are communication parameters with the communication key size $d^{K_\text{comm}}\in\mathbb{N}$.

\Cref{eq:third_layer} is similar to~\citet{RIMs}. 
We only update the top-$k$ selected mechanisms from the option selection module, but through this transfer of information, each active mechanism can update its internal belief state and have improved exploration through contextualization. 
We note that there are many choices in what components to share across options. We explore these choices and their implications in our empirical analysis.

\paragraph{Fully Connected Module.} 
Lastly, we have a separate fully connected layer for each option-specific intra-option $\pi_{\omega}$ and termination policy $\beta_{\omega}$, which take $b_{\omega}$ as input. 
The intra-option policy provides the final output action $a$, and the termination determines whether $\omega$ should terminate. 

\subsection{Implementation} We first describe our choice for representing factored belief states. 
Under the state uniformity assumption discussed in \citet{non_markov_process}, we assume the optimal policy network based on the agent's history is equivalent to the network conditioned on the true state space. Hence, we refer to the representation learned as an implicit belief state. More explicit methods for modeling belief states have been considered, for example, as outlined in \citet{igl2018deep}. While the CRADOL framework is certainly compatible with this kind of explicit belief state modeling, we have chosen to implement the belief state implicitly and denote it as the factored belief state in order to have a fair empirical comparison to the RIMs method \cite{RIMs} that we build off. 

Our implementation draws inspiration from soft-actor critic~\cite{haarnoja2018soft} to enable sample efficient off-policy optimization of the option-value functions, intra-option policy, and beta policy. 
In the appendix, we describe the implementation of our approach (see \cref{alg:cradol}) and additional details about the architecture.

\section{Related Work} \label{sec:related_work}

\paragraph{Hierarchical RL.}
There have been two major high-level approaches in the recent literature for achieving efficient HRL: the options framework to learn abstract skills \cite{sutton1991,bacon2016optioncritic,omidshafiei18casl,riemer18options,riemer2020role} and goal-conditioned hierarchies to learn abstract sub-goals~\cite{nachum2019nearoptimal, Levy2017HierarchicalA,kulkarni,kim20hmat}. 
Goal-conditioned HRL approaches require a pre-defined representation of the environment and mapping of observation space to goal space, whereas the OC framework facilitates long-timescale credit assignment by dividing the problem into pieces and learning higher-level skills. 
Our work is most similar to \citet{khetarpal_options_interest} which learns a smaller initiation set through an attention mechanism to better focus an option to a region of the state space, and hence achieve specialization of skills. 
However, it does not leverage the focus to abstract and minimize the size of the belief space as CRADOL does. 
We consider a more explicit method of specialization by leveraging context-specific independence for representation abstraction. 
Our approach could also potentially consider learning initiation sets, so we consider our contributions to be orthogonal. 

While OC provides a temporal decomposition of the problem, other approaches such as Feudal Networks \cite{feudalRL} decompose problems with respect to the state space. Feudal approaches use a manager to learn more abstract goals over a longer temporal scale and worker modules to perform more primitive actions with fine temporal resolution. Our approach employs a combination of both visions, necessitating both temporal and state abstraction for effective decomposition. Although some approaches such as the MAXQ framework \cite{maxq} employ both, they involve learning recursively optimal policies that can be highly sub-optimal \cite{BerliacHierachialRL2019}. 

\begin{figure}[t!]
	\centering
	\includegraphics[width=1\linewidth]{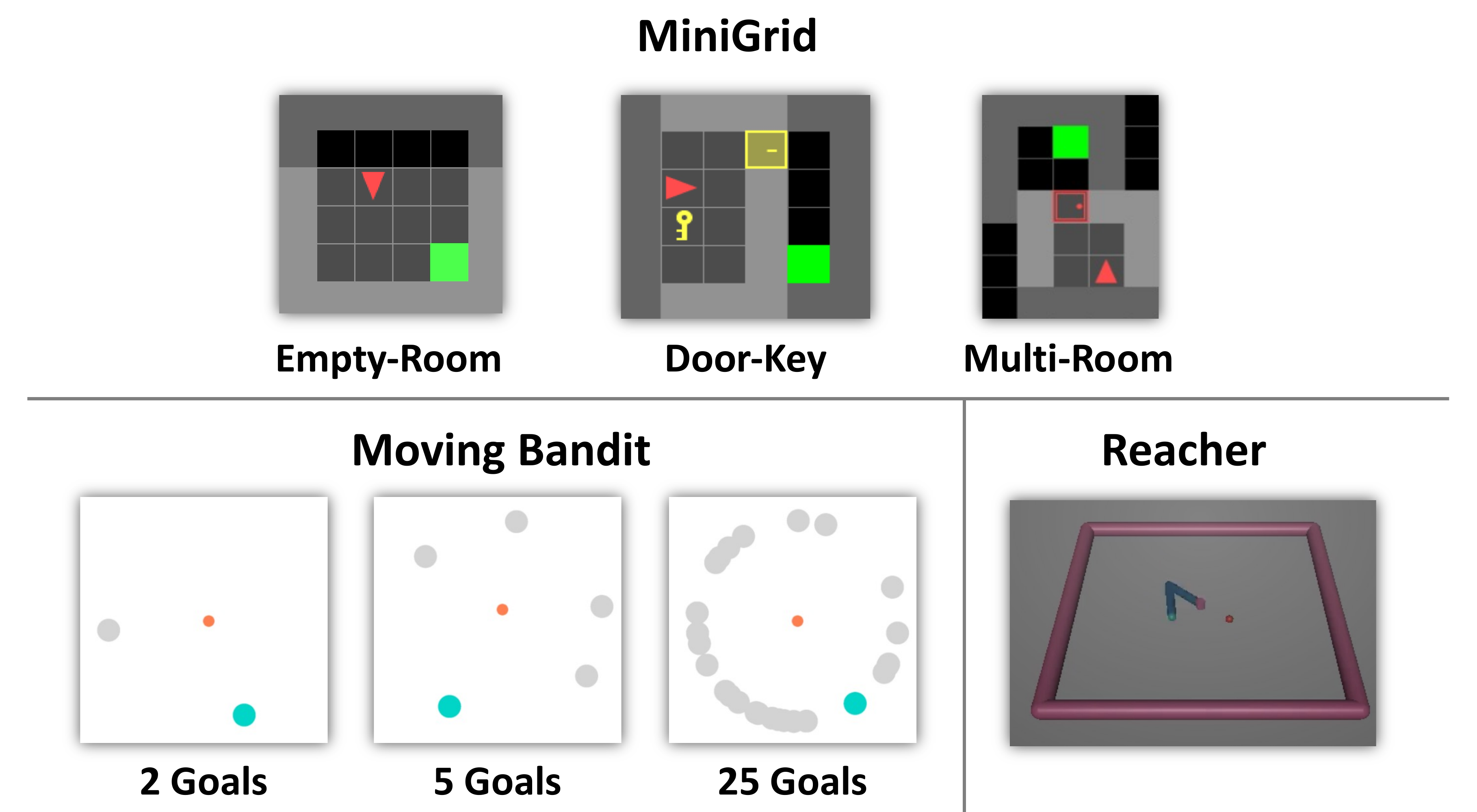}
	\caption{Visualization of MiniGrid~\cite{gym_minigrid}, Moving Bandit~\cite{moving_bandit}, and Reacher~\cite{brockman2016openai} domains.}
	\label{fig:domains}
\end{figure}

\paragraph{State Abstractions.}
Our approach is related to prior work that considers the importance of state abstraction and the decomposition of the learning problem \cite{KONIDARIS20191, skill_learning_abs, abs_irrelevant}. 
Notable methods of state abstraction include PAC state abstraction, which achieves correct clustering with high probability with respect to a distribution over learning problems \cite{pmlr-v80-abel18a}. 
This abstraction method can have limited applicability to deep RL methods such as CRADOL.
\citet{casual_state_rep} has been able to learn task agnostic state abstractions by identifying casual states in the POMDP setting, whereas our approach considers discovering abstractions for sub-task specific learning. 
\citet{abstraction_kaelbling} introduces the importance of abstraction in planning problems with \citet{chitnis2020camps} performing a context-specific abstraction for the purposes of decomposition in planning-related tasks. CRADOL extends this work by exploring context-specific abstraction in HRL. 

\section{Evaluation}\label{sec:evaluation}
We demonstrate CRADOL’s efficacy on a diverse suite of domains. 
The code is available at 
\url{https://git.io/JucVH} and the videos are available at 
\url{https://bit.ly/3tpJc8Z}. We explain further details on experimental settings, including domains and hyperparameters, in the appendix. 

\begin{table*}[t!]
	\centering
	\renewrobustcmd{\bfseries}{\fontseries{b}\selectfont}
	\sisetup{detect-weight,mode=text,group-minimum-digits = 4}
	{\small
		\tabcolsep=0.10cm
		\begin{tabular}[t]{l
				S[separate-uncertainty,table-figures-uncertainty=1,table-format=2.2(2)]
				S[separate-uncertainty,table-figures-uncertainty=1,table-format=3(3)]
				S[separate-uncertainty,table-figures-uncertainty=1,table-format=2.2(2)]
				S[separate-uncertainty,table-figures-uncertainty=1,table-format=3(3)]
				S[separate-uncertainty,table-figures-uncertainty=1,table-format=2.2(2)]
				S[separate-uncertainty,table-figures-uncertainty=1,table-format=3(3)]
				S[separate-uncertainty,table-figures-uncertainty=1,table-format=2.2(2)]
				S[separate-uncertainty,table-figures-uncertainty=1,table-format=3(3)]
			}
			\toprule
			Algorithm  & \multicolumn{2}{c}{Abstraction} & \multicolumn{2}{c}{MiniGrid Empty} & \multicolumn{2}{c}{MiniGrid MultiRoom} & \multicolumn{2}{c}{MiniGrid KeyDoor}         \\
			\cmidrule(lr){2-3} \cmidrule(lr){4-5} \cmidrule(lr){6-7} \cmidrule(lr){8-9} & \text{Temporal} & \text{State} & $\bar{V}$ & AUC & $\bar{V}$ & AUC  & $\bar{V}$ & AUC                 \\ \midrule
			A2C &  \ding{55}  &  \ding{55} &  0.71 \pm 0.25 & 364 \pm 54 & \bfseries 0.62 \pm 0.26 & 104 \pm 27  & 0.04 \pm 0.07 & 16 \pm 10    \\
			SAC &  \ding{55}  &  \ding{55} &  0.65 \pm 0.29 & 350 \pm 124 & \bfseries 0.62 \pm 0.09 & 162 \pm 39  & 0.48 \pm 0.42 & 286 \pm 138    \\
			OC &  \ding{51}  &  \ding{55} & 0.56 \pm 0.36 & 312\pm 143  & 0.40 \pm 0.15 & 108\pm 45  & 0.55 \pm 0.44  & 252 \pm 217  \\
			A2C-RIM &  \ding{55}  &  \ding{51} & 0.57 \pm 0.40 & 283 \pm 47 & 0.19 \pm 0.23  & 52 \pm 22 & 0.57 \pm 0.40  & 12 \pm 1  \\
			\midrule
			CRADOL    &  \ding{51}  &  \ding{51} & \bfseries 0.92 \pm 0.12 & \bfseries 470 \pm 21 & \bfseries 0.75 \pm 0.06 & \bfseries 200 \pm 27 & \bfseries 0.87 \pm 0.17 & \bfseries 499 \pm 25  \\
			\bottomrule
		\end{tabular}
	}
		\caption{$\bar{V}$ and Area under the Curve (AUC) in MiniGrid domains. Table shows mean and standard deviation computed with 10 random seeds. Best results in bold (computed by $t$-test with $p < 0.05$). Note that CRADOL has the highest $\bar{V}$ and AUC compared to non-HRL (A2C, SAC), HRL (OC), and modular recurrent neural network (A2C-RIM) baselines. Figures 8-11 in the appendix show number of steps taken to converge for these experiments.}
		\label{table:results_comparison} 
\end{table*}

\subsection{Experimental Setup}
\paragraph{Domains.}
We demonstrate the performance of our approach with domains shown in \cref{fig:domains}. 
MiniGrid domains are well suited for hierarchical learning due to the natural emergence of skills needed to accomplish the goal. Moving Bandits considers the performance of CRADOL with extraneous features in sparse reward settings. Lastly, the Reacher domain observes the effects of CRADOL on low-level observation representations.

\begin{itemize}
    \itemsep 0pt
    \item \textbf{MiniGrid}~\cite{gym_minigrid}: A library of open-source grid-world domains in sparse reward settings with image observations. Each grid contains exactly zero or one object with possible object types such as the wall, door, key, ball, box, and goal indicated by different colors. The goal for the domain can vary from obtaining a key to matching similar colored objects. The agent receives a sparse reward of 1 when it successfully reaches the green goal tile, and 0 for failure. 
    \item \textbf{Moving Bandit}~\cite{moving_bandit}: This 2D sparse reward setting considers a number of marked positions in the environment that change at random at every episode, with 1 of the positions being the correct goal. An agent receives a reward of 1 and terminates when the agent is close to the correct goal position, and receives 0 otherwise.        
    \item \textbf{Reacher}~\cite{brockman2016openai}: In this simulated MuJoCo task of OpenAI Gym environment, a robot arm consisting of 2 linkages with equal length must reach a random red target placed randomly at the beginning of each episode. We modify the domain to be a sparse reward setting: the agent receives a reward signal of 1 when its euclidean distance to the target is within a threshold, and 0 otherwise. 
\end{itemize}

\paragraph{Baselines.}
We compare CRADOL to the following non-hierarchical, hierarchical, and modular recurrent neural network baselines:
\begin{itemize}
    \itemsep 0pt
    \item \textbf{A2C}~\cite{a2c}:
    This on-policy method considers neither context-specific nor temporal abstraction. 
    \item \textbf{SAC}~\cite{haarnoja2018soft}: This entropy-maximization off-policy method considers neither context-specific nor temporal abstraction. 
    \item \textbf{OC}~\cite{bacon2016optioncritic}: 
    We consider an off-policy implementation of OC based on the SAC method to demonstrate the performance of a hierarchical method considering only temporal abstraction.
    \item \textbf{A2C-RIM}~\cite{RIMs}. This method considers A2C with recurrent independent mechanisms, a baseline that allows us to observe the performance of a method employing context-specific abstraction only.
\end{itemize}

\subsection{Results}

\begin{figure}[tb]
\centering
\begin{subfigure}[t]{\linewidth}
  \includegraphics[width=0.9\linewidth]{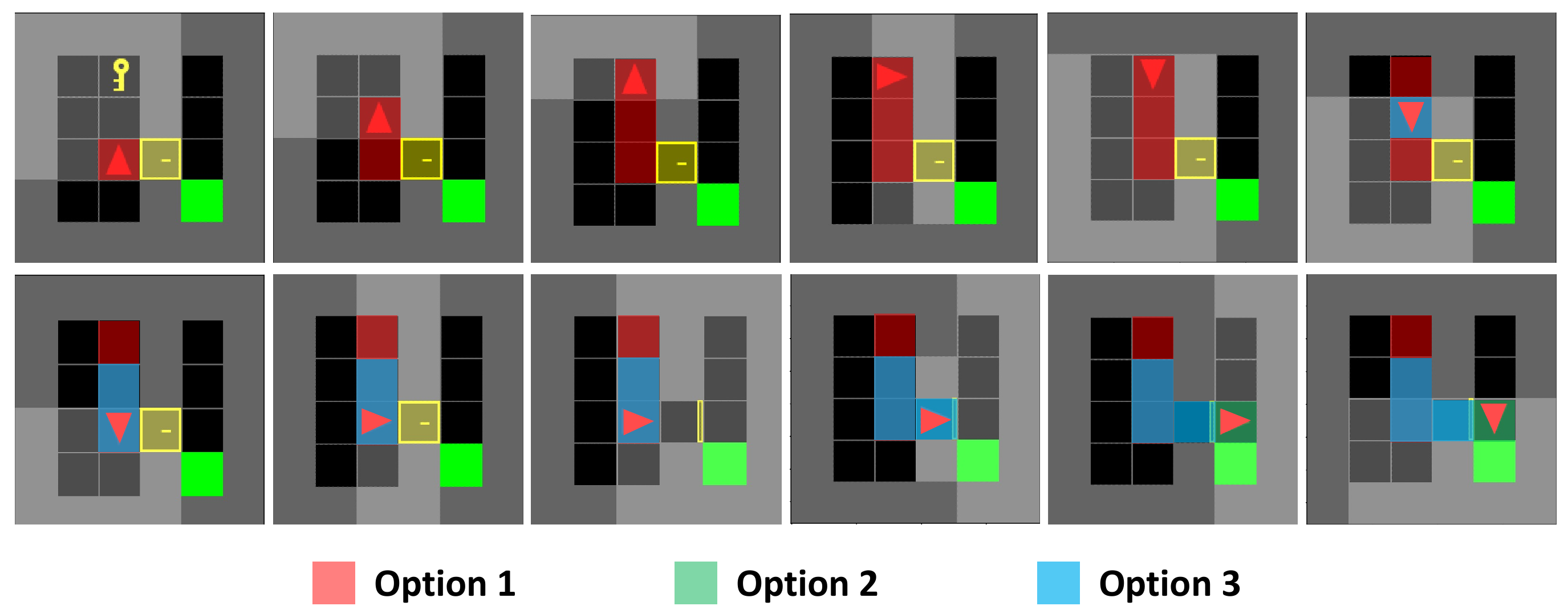}
  \caption{Option Trajectory}
  \label{fig:option-trajectory} 
\end{subfigure}
\begin{subfigure}[b]{0.465\linewidth}
  \includegraphics[width=1.0\linewidth,page=1]{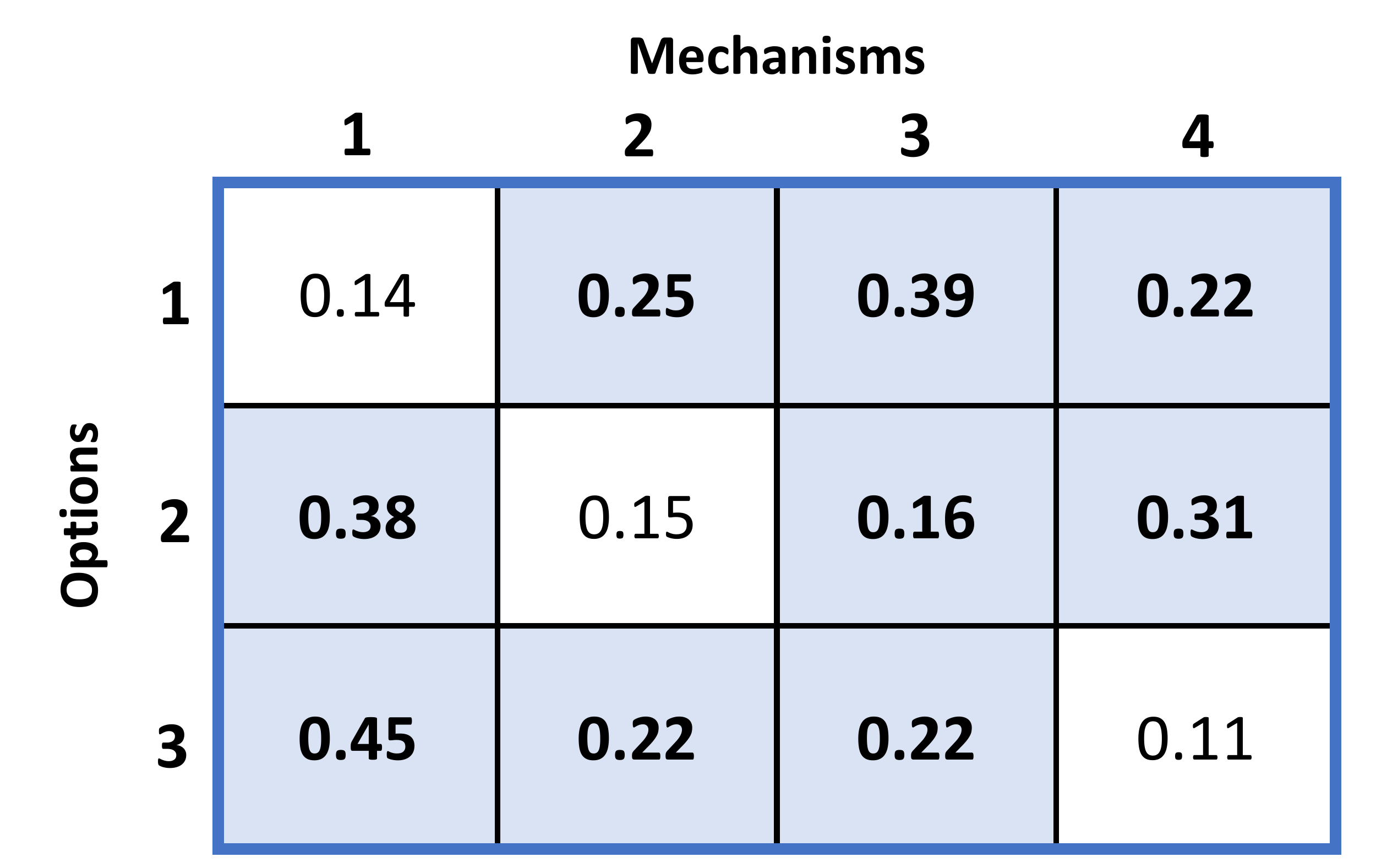}
  \caption{Mechanisms Selection}
  \label{fig:look-up}
\end{subfigure}
\hfill
\begin{subfigure}[b]{0.465\linewidth}
  \includegraphics[width=1.0\linewidth,page=2]{figs/lookup_table.pdf}
  \caption{Option Correlation}
  \label{fig:correlation}
\end{subfigure}
\caption{(a) Temporal abstraction by the use of 3 options for following tasks: option 1 for getting the key, option 2 for opening the door, and option 3 for going to the door. (b) Each option maps to a unique subset of mechanisms, corresponding to their unique functions in the domain. (c) There is low correlation between option 1 \& 2 and option 2 \& 3 but higher correlation between option 1 \& 2 corresponding to the shared belief states between them.}
\end{figure}

\begin{question}
Does context-specific abstraction help achieve sample efficiency?
\end{question}
\noindent To answer this question, we compare performance in the MiniGrid domains.
With a sparse reward function and dense belief state representation (i.e., image observation), MiniGrid provides the opportunity to test the temporal and context-specific abstraction capabilities of our method. 
\cref{table:results_comparison} shows both final task-level performance ($\bar{V}$) (i.e., final episodic average reward measured at the end of learning) and area under the learning curve (AUC). Higher values indicate better results for both metrics. 

Overall, for all three scenarios of MiniGrid, CRADOL has the highest $\bar{V}$ and AUC than the baselines. 
We observe that OC has a lower performance than CRADOL due to the inability of the options learned to diversify by considering the entire belief state space and the high termination probability of each option. 
Both A2C and SAC result in sub-optimal performance due to their failure in sparse reward settings. 
Finally, due to inefficient exploration and large training time required for A2C-RIM to converge, it is unable to match the performance of CRADOL. 
We see a smaller difference between CRADOL and these baselines for the Empty domain, as there is a smaller amount of context-specific abstraction that is required in this simplest setting of the MiniGrid domain. 
For the Multi-Room domain which is more difficult than the EmptyRoom domain, there is an increasingly larger gap as the agent needs to consider only the belief states in one room when trying to exit that room and belief states of the other room when trying to reach the green goal. 
Lastly, we see the most abstraction required for the Key Door domain where the baselines are unable to match the performance of CRADOL. As described in \cref{fig:motivation}, OC equipped with only temporal abstraction is unable to match the performance of CRADOL consisting of both temporal and context-specific abstraction.

\begin{question}
What does it mean to have diverse options?
\end{question}
\noindent We visualize the behaviors of options for temporal abstraction and mechanisms for context-specific abstraction to further understand whether options are able to learn diverse sub-policies. 
\cref{fig:option-trajectory} shows the option trajectory in the DoorKey domain for the following (learned) sub-tasks: getting the key, opening the door, and going to the door. 
We find that each option is only activated for one sub-task. 
\cref{fig:look-up} shows the mapping between options and mechanisms, and we see that each option maps to a unique subset of mechanisms. To understand whether these mechanisms have mapped to different factors of the belief state (and hence have diverse parameters), \cref{fig:correlation} computes the correlation between options, measured by the Pearson product-moment correlation method \cite{freedman2007statistics}. 
We find low correlation between option 1 \& 2 and option 2 \& 3 but higher correlation between option 1 \& 3. 
Specifically, we observe a high correlation between option 1 \& 3 in getting the key and opening the door due to the shared states between them, because opening the door is highly dependent on obtaining the key in the environment.
This visualization empirically verifies our hypothesis that both temporal and context-specific abstraction are necessary to learn diverse and complementary option policies. 

\begin{figure}[t]
  \begin{center}
    \includegraphics[height=0.5\linewidth]{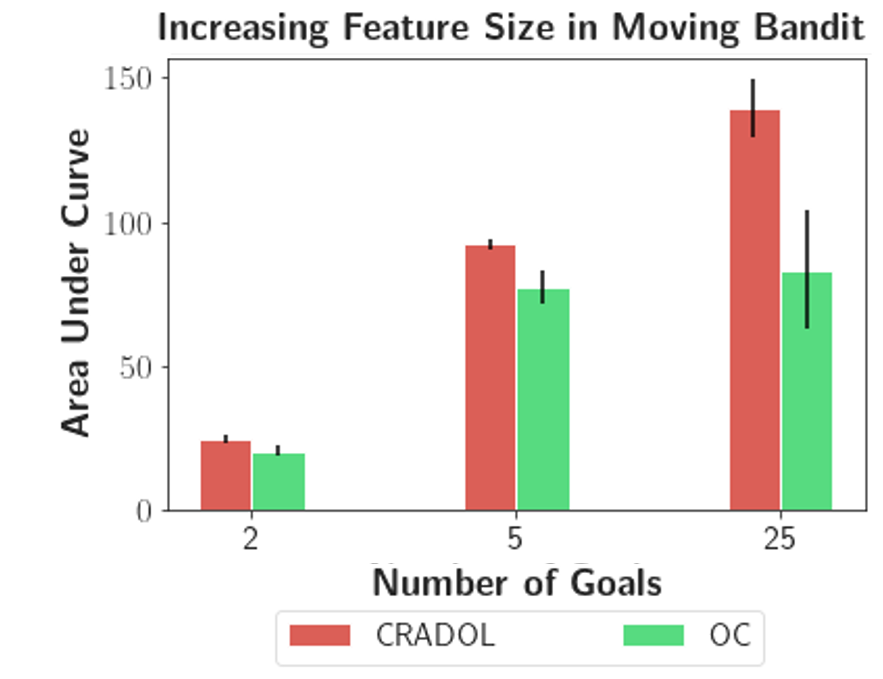}
   \caption{AUC between CRADOL and OC in Moving Bandit domain. As the number of spurious features on the x-axis increases, the gap between the AUC performance between CRADOL and OC increases. This indicates a greater need for context-specific abstraction. Note that we see a different AUC scale across the number of goals simply due to different max train iteration for each goal setting. Mean and 95\% confidence interval computed for 10 seeds are shown.}
   \label{fig:auc_bandit}
   \end{center}
\end{figure}

\begin{question}
How does performance change as the need for context-specific abstraction increases?
\end{question}
\noindent In order to understand the full benefits of context-specific abstraction, we observe the performance with increasing context-specific abstraction in the Moving Bandits domain. 
We consider the performance determined by AUC for an increasing number of spurious features in the observation.
Namely, we add 3 \& 23 additional goals to the original 2 goal observation to test the capabilities of CRADOL (see \cref{fig:domains}). Increasing the number of goals requires an increasing amount of context-specific abstraction, as there are more spurious features the agent must ignore to learn to move to the correct goal location as indicated in its observation. 
As shown in \cref{fig:auc_bandit}, CRADOL performs significantly better than OC as the number of spurious goal locations it must ignore increases. 
We expect that this result is due to the CRADOL's capability of both temporal and context-specific abstraction.
\begin{figure}[t]
  \begin{center}
  \centerline{\includegraphics[width=0.95\linewidth]{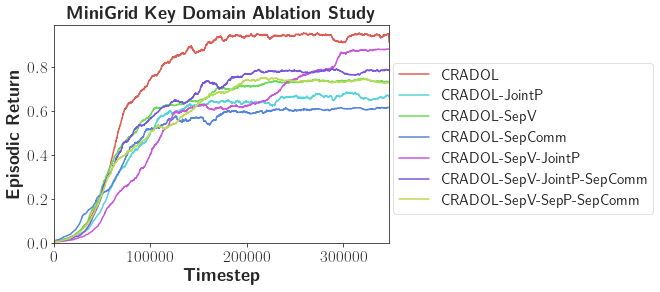}}
  \caption{Ablation study experimenting with various components that can be shared and not shared between options in the option learning process shown in \cref{fig:approach}. Too much sharing or too little sharing between options can lead to sub-optimal performance due to a lack of coordination. This result shows the mean computed for 10 seeds.} 
   \label{fig:ablation}
   \end{center}
    \vskip-0.18in
\end{figure}

\begin{question} What is shared between options in the context of mechanisms?
\end{question}
\noindent As described in \cref{fig:approach}, there are 4 components that make up the option learning model. 
In order to investigate which are necessary to share between options, we perform an ablation study with a subset of the available choices shown in \cref{fig:ablation}.
Specifically, we study sharing of the look-up table $W^{\text{table}}$ in the input attention module between options (CRADOL-JointP), learning a separate parameter $W^{\text{value}}$ between options (CRADOL-SepV), learning separate parameters for ($W^{Q_{\text{comm}}}, W^{K_{\text{comm}}}$, $W^{V_{\text{comm}}}$) of the sparse communication layer for each option (CRADOL-SepComm), and three other combinations of these three modules.

We find the lowest performance for the method with a separate sparse communication module for each option. 
We hypothesize that this is due to a lack of coordination between each option in updating their active mechanisms and ensuring other non-active mechanisms are learning separate and complementary parameters. 
Having a joint look-up table $W^{\text{table}}$ results in the second-lowest performance. This effectively maps each option to the same set of mechanisms, leading to a lack of diversity between option policies and only allowing for the difference in the fully connected layer of each option. 
Lastly, we observe the third-lowest performance with a separate parameter $W^{\text{value}}$ between options. Each option learning from a different representation can lead to similar option-policies and factored belief states for certain sub-goals unaccounted for. Other combinations reaffirm that sharing the sparse communication larger is essential for coordinated behavior when learning option policies. 

\begin{figure}[t]
  \begin{center}
  \centerline{\includegraphics[height=0.5\linewidth]{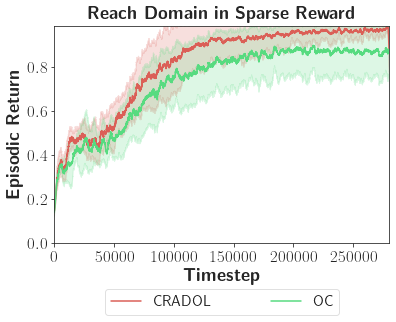}}
   \caption{In domains with small or negligible context-specific abstraction, the benefit of CRADOL is not as significant compared to the Reacher domain, where the low-dimensional observation representation does not require context-specific abstraction.  
   This figure shows the mean and 95\% confidence interval computed for 10 seeds.}
   \label{fig:reach}
   \end{center}
   \vskip-0.18in
\end{figure}

\begin{question}
Is context-specific abstraction always beneficial? 
\end{question}
\noindent The Reach domain allows us to observe the effects of our method when there is little benefit of reducing the problem size, namely, when the observation is incredibly small. This domain does not require context-specific abstraction as the entire belief state space consists of relevant information to achieve the goal at hand. Specifically, the low-level representation of the observation as a 4-element observation vector, with the first 2 elements containing the generalized positions and the next 2 elements containing the generalized velocities of the two arm joints, are essential to reach the goal location. As expected in \cref{fig:reach}, the performance between CRADOL and OC is similar in this domain as the observation space does not contain any features that are useful for CRADOL to perform context-specific abstraction. We note our gains are larger for problems where they have larger intrinsic dimensionality. 
 
\section{Conclusion} \label{sec:conclusion}
In this paper, we have introduced Context-Specific Representation Abstraction for Deep Option Learning (CRADOL) for incorporating both temporal and state abstraction for the effective decomposition of a problem into simpler components. The key idea underlying our proposed formulation is to map each option to a reduced state space, effectively considering each option as a subset of mechanisms. We evaluated our method on several benchmarks with varying levels of required abstraction in sparse reward settings. Our results indicate that CRADOL is able to decompose the problems at hand more effectively than state-of-the-art HRL, non-HRL, and modular recurrent neural network baselines. We hope that our work can help provide the community with a theoretical foundation to build off for addressing the deficiencies in HRL methods. 
\newpage
\section*{Acknowledgements}
Research funded by IBM, Samsung (as part of the MIT-IBM Watson AI Lab initiative) and computational support through Amazon Web Services.

\bibliography{aaai22}

\begin{thebibliography}{39}
\providecommand{\natexlab}[1]{#1}

\bibitem[{Abel et~al.(2018)Abel, Arumugam, Lehnert, and
  Littman}]{pmlr-v80-abel18a}
Abel, D.; Arumugam, D.; Lehnert, L.; and Littman, M. 2018.
\newblock State Abstractions for Lifelong Reinforcement Learning.
\newblock In \emph{Proceedings of the 35th International Conference on Machine
  Learning}, volume~80 of \emph{Proceedings of Machine Learning Research},
  10--19. PMLR.

\bibitem[{Bacon, Harb, and Precup(2017)}]{bacon2016optioncritic}
Bacon, P.-L.; Harb, J.; and Precup, D. 2017.
\newblock The Option-Critic Architecture.
\newblock In \emph{Proceedings of the Thirty-First AAAI Conference on
  Artificial Intelligence}, AAAI'17, 1726–1734. AAAI Press.

\bibitem[{Boutilier et~al.(1996)Boutilier, Friedman, Goldszmidt, and
  Koller}]{context_definition}
Boutilier, C.; Friedman, N.; Goldszmidt, M.; and Koller, D. 1996.
\newblock Context-Specific Independence in Bayesian Networks.
\newblock In \emph{Proceedings of the Twelfth International Conference on
  Uncertainty in Artificial Intelligence}, UAI'96, 115–123. Morgan Kaufmann
  Publishers Inc.

\bibitem[{Brockman et~al.(2016)Brockman, Cheung, Pettersson, Schneider,
  Schulman, Tang, and Zaremba}]{brockman2016openai}
Brockman, G.; Cheung, V.; Pettersson, L.; Schneider, J.; Schulman, J.; Tang,
  J.; and Zaremba, W. 2016.
\newblock OpenAI Gym.
\newblock \emph{CoRR}, abs/1606.01540.

\bibitem[{Chevalier-Boisvert, Willems, and Pal(2018)}]{gym_minigrid}
Chevalier-Boisvert, M.; Willems, L.; and Pal, S. 2018.
\newblock Minimalistic Gridworld Environment for OpenAI Gym.
\newblock \url{https://github.com/maximecb/gym-minigrid}.

\bibitem[{Chitnis et~al.(2020)Chitnis, Silver, Kim, Kaelbling, and
  Lozano{-}P{\'{e}}rez}]{chitnis2020camps}
Chitnis, R.; Silver, T.; Kim, B.; Kaelbling, L.~P.; and Lozano{-}P{\'{e}}rez,
  T. 2020.
\newblock CAMPs: Learning Context-Specific Abstractions for Efficient Planning
  in Factored MDPs.
\newblock \emph{CoRR}, abs/2007.13202.

\bibitem[{Daniel et~al.(2016)Daniel, Van~Hoof, Peters, and
  Neumann}]{determine_options_daniel}
Daniel, C.; Van~Hoof, H.; Peters, J.; and Neumann, G. 2016.
\newblock Probabilistic Inference for Determining Options in Reinforcement
  Learning.
\newblock 104(2–3).

\bibitem[{Dietterich(2000)}]{maxq}
Dietterich, T. 2000.
\newblock Hierarchical Reinforcement Learning with the MAXQ Value Function
  Decomposition.
\newblock \emph{The Journal of Artificial Intelligence Research (JAIR)}, 13.

\bibitem[{Flet-Berliac(2019)}]{BerliacHierachialRL2019}
Flet-Berliac, Y. 2019.
\newblock The Promise of Hierarchical Reinforcement Learning.
\newblock \emph{The Gradient}.

\bibitem[{Frans et~al.(2017)Frans, Ho, Chen, Abbeel, and
  Schulman}]{moving_bandit}
Frans, K.; Ho, J.; Chen, X.; Abbeel, P.; and Schulman, J. 2017.
\newblock Meta Learning Shared Hierarchies.
\newblock \emph{CoRR}, abs/1710.09767.

\bibitem[{Freedman, Pisani, and Purves(2007)}]{freedman2007statistics}
Freedman, D.; Pisani, R.; and Purves, R. 2007.
\newblock Statistics (international student edition).
\newblock \emph{Pisani, R. Purves, 4th edn. WW Norton \& Company, New York}.

\bibitem[{Goyal et~al.(2021)Goyal, Lamb, Hoffmann, Sodhani, Levine, Bengio, and
  Sch{\"o}lkopf}]{RIMs}
Goyal, A.; Lamb, A.; Hoffmann, J.; Sodhani, S.; Levine, S.; Bengio, Y.; and
  Sch{\"o}lkopf, B. 2021.
\newblock Recurrent Independent Mechanisms.
\newblock In \emph{International Conference on Learning Representations}.

\bibitem[{Guestrin et~al.(2003{\natexlab{a}})Guestrin, Koller, Parr, and
  Venkataraman}]{Guestrin_2003}
Guestrin, C.; Koller, D.; Parr, R.; and Venkataraman, S. 2003{\natexlab{a}}.
\newblock Efficient Solution Algorithms for Factored MDPs.
\newblock \emph{Journal of Artificial Intelligence Research}, 19: 399–468.

\bibitem[{Guestrin et~al.(2003{\natexlab{b}})Guestrin, Koller, Parr, and
  Venkataraman}]{factoredmdps}
Guestrin, C.; Koller, D.; Parr, R.; and Venkataraman, S. 2003{\natexlab{b}}.
\newblock Efficient solution algorithms for factored MDPs.
\newblock \emph{Journal of Artificial Intelligence Research}, 19: 399--468.

\bibitem[{Haarnoja et~al.(2018)Haarnoja, Zhou, Abbeel, and
  Levine}]{haarnoja2018soft}
Haarnoja, T.; Zhou, A.; Abbeel, P.; and Levine, S. 2018.
\newblock Soft Actor-Critic: Off-Policy Maximum Entropy Deep Reinforcement
  Learning with a Stochastic Actor.
\newblock In Dy, J.; and Krause, A., eds., \emph{Proceedings of the 35th
  International Conference on Machine Learning}, volume~80 of \emph{Proceedings
  of Machine Learning Research}, 1861--1870. PMLR.

\bibitem[{Igl et~al.(2018)Igl, Zintgraf, Le, Wood, and Whiteson}]{igl2018deep}
Igl, M.; Zintgraf, L.; Le, T.~A.; Wood, F.; and Whiteson, S. 2018.
\newblock Deep Variational Reinforcement Learning for {POMDP}s.
\newblock In \emph{Proceedings of the 35th International Conference on Machine
  Learning}, volume~80 of \emph{Proceedings of Machine Learning Research},
  2117--2126. PMLR.

\bibitem[{Jong and Stone(2005)}]{abs_irrelevant}
Jong, N.~K.; and Stone, P. 2005.
\newblock State Abstraction Discovery from Irrelevant State Variables.
\newblock In \emph{Proceedings of the 19th International Joint Conference on
  Artificial Intelligence}, IJCAI'05, 752–757. Morgan Kaufmann Publishers
  Inc.

\bibitem[{Kaelbling, Littman, and Cassandra(1998)}]{POMDPs}
Kaelbling, L.~P.; Littman, M.~L.; and Cassandra, A.~R. 1998.
\newblock Planning and acting in partially observable stochastic domains.
\newblock \emph{Artificial intelligence}, 101(1-2): 99--134.

\bibitem[{Kamat and Precup(2020)}]{diversity}
Kamat, A.; and Precup, D. 2020.
\newblock Diversity-Enriched Option-Critic.
\newblock \emph{CoRR}, abs/2011.02565.

\bibitem[{Khetarpal et~al.(2020)Khetarpal, Klissarov, Chevalier-Boisvert,
  Bacon, and Precup}]{khetarpal_options_interest}
Khetarpal, K.; Klissarov, M.; Chevalier-Boisvert, M.; Bacon, P.-L.; and Precup,
  D. 2020.
\newblock Options of interest: Temporal abstraction with interest functions.
\newblock In \emph{Proceedings of the AAAI Conference on Artificial
  Intelligence}, volume~34, 4444--4451.

\bibitem[{Kim et~al.(2019)Kim, Liu, Omidshafiei, Lopez{-}Cot, Riemer, Habibi,
  Tesauro, Mourad, Campbell, and How}]{kim20hmat}
Kim, D.; Liu, M.; Omidshafiei, S.; Lopez{-}Cot, S.; Riemer, M.; Habibi, G.;
  Tesauro, G.; Mourad, S.; Campbell, M.; and How, J.~P. 2019.
\newblock Learning Hierarchical Teaching in Cooperative Multiagent
  Reinforcement Learning.
\newblock \emph{CoRR}, abs/1903.03216.

\bibitem[{Konidaris(2019)}]{KONIDARIS20191}
Konidaris, G. 2019.
\newblock On the necessity of abstraction.
\newblock \emph{Current Opinion in Behavioral Sciences}, 29: 1--7.
\newblock Artificial Intelligence.

\bibitem[{Konidaris and Barto(2009)}]{skill_learning_abs}
Konidaris, G.; and Barto, A. 2009.
\newblock Efficient Skill Learning Using Abstraction Selection.
\newblock In \emph{Proceedings of the 21st International Jont Conference on
  Artifical Intelligence}, IJCAI'09, 1107–1112. Morgan Kaufmann Publishers
  Inc.

\bibitem[{Konidaris, Kaelbling, and Lozano-Perez(2018)}]{abstraction_kaelbling}
Konidaris, G.; Kaelbling, L.~P.; and Lozano-Perez, T. 2018.
\newblock From Skills to Symbols: Learning Symbolic Representations for
  Abstract High-Level Planning.
\newblock \emph{J. Artif. Int. Res.}, 215–289.

\bibitem[{Kulkarni and Narasimhan(2016)}]{kulkarni}
Kulkarni, T.~D.; and Narasimhan, K.~R. 2016.
\newblock Hierarchical Deep Reinforcement Learning: Integrating Temporal
  Abstraction and Intrinsic Motivation.
\newblock \emph{Neural Information Processing Systems 2016}.

\bibitem[{Levy, Platt, and Saenko(2019)}]{Levy2017HierarchicalA}
Levy, A.; Platt, R.; and Saenko, K. 2019.
\newblock Hierarchical Reinforcement Learning with Hindsight.
\newblock In \emph{International Conference on Learning Representations}.

\bibitem[{Majeed and Hutter(2018)}]{non_markov_process}
Majeed, S.~J.; and Hutter, M. 2018.
\newblock On Q-Learning Convergence for Non-Markov Decision Processes.
\newblock In \emph{Proceedings of the 27th International Joint Conference on
  Artificial Intelligence}, IJCAI'18, 2546–2552. AAAI Press.
\newblock ISBN 9780999241127.

\bibitem[{Mankowitz, Mann, and Mannor(2016)}]{Mankowitz2016AdaptiveSA}
Mankowitz, D.~J.; Mann, T.~A.; and Mannor, S. 2016.
\newblock Adaptive Skills Adaptive Partitions (ASAP).
\newblock In \emph{Advances in Neural Information Processing Systems},
  volume~29. Curran Associates, Inc.

\bibitem[{Mnih et~al.(2016)Mnih, Badia, Mirza, Graves, Lillicrap, Harley,
  Silver, and Kavukcuoglu}]{a2c}
Mnih, V.; Badia, A.~P.; Mirza, M.; Graves, A.; Lillicrap, T.; Harley, T.;
  Silver, D.; and Kavukcuoglu, K. 2016.
\newblock Asynchronous Methods for Deep Reinforcement Learning.
\newblock In \emph{Proceedings of The 33rd International Conference on Machine
  Learning}, volume~48 of \emph{Proceedings of Machine Learning Research},
  1928--1937. PMLR.

\bibitem[{Nachum et~al.(2019)Nachum, Gu, Lee, and
  Levine}]{nachum2019nearoptimal}
Nachum, O.; Gu, S.; Lee, H.; and Levine, S. 2019.
\newblock Near-Optimal Representation Learning for Hierarchical Reinforcement
  Learning.
\newblock In \emph{International Conference on Learning Representations}.

\bibitem[{Omidshafiei et~al.(2018)Omidshafiei, Kim, Pazis, and
  How}]{omidshafiei18casl}
Omidshafiei, S.; Kim, D.-K.; Pazis, J.; and How, J.~P. 2018.
\newblock Crossmodal Attentive Skill Learner.
\newblock In \emph{Proceedings of the 17th International Conference on
  Autonomous Agents and MultiAgent Systems}, AAMAS '18, 139–146.
  International Foundation for Autonomous Agents and Multiagent Systems.

\bibitem[{Precup and Sutton(2000)}]{doina_thesis}
Precup, D.; and Sutton, R.~S. 2000.
\newblock \emph{Temporal Abstraction in Reinforcement Learning}.
\newblock Ph.D. thesis, University of Massachusetts Amherst.
\newblock AAI9978540.

\bibitem[{Puterman(1994)}]{smdp}
Puterman, M.~L. 1994.
\newblock \emph{Markov Decision Processes: Discrete Stochastic Dynamic
  Programming}.
\newblock USA: John Wiley \& Sons, Inc., 1st edition.
\newblock ISBN 0471619779.

\bibitem[{Riemer et~al.(2020)Riemer, Cases, Rosenbaum, Liu, and
  Tesauro}]{riemer2020role}
Riemer, M.; Cases, I.; Rosenbaum, C.; Liu, M.; and Tesauro, G. 2020.
\newblock On the role of weight sharing during deep option learning.
\newblock In \emph{Proceedings of the AAAI Conference on Artificial
  Intelligence}, volume~34, 5519--5526.

\bibitem[{Riemer, Liu, and Tesauro(2018)}]{riemer18options}
Riemer, M.; Liu, M.; and Tesauro, G. 2018.
\newblock Learning Abstract Options.
\newblock In Bengio, S.; Wallach, H.; Larochelle, H.; Grauman, K.;
  Cesa-Bianchi, N.; and Garnett, R., eds., \emph{Advances in Neural Information
  Processing Systems}, volume~31. Curran Associates, Inc.

\bibitem[{Sutton, Precup, and Singh(1999)}]{sutton1991}
Sutton, R.~S.; Precup, D.; and Singh, S. 1999.
\newblock Between MDPs and semi-MDPs: A framework for temporal abstraction in
  reinforcement learning.
\newblock \emph{Artificial Intelligence}, 112(1): 181--211.

\bibitem[{Vezhnevets et~al.(2016)Vezhnevets, Mnih, Osindero, Graves, Vinyals,
  Agapiou, and kavukcuoglu}]{NIPS2016_c4492cbe}
Vezhnevets, A.; Mnih, V.; Osindero, S.; Graves, A.; Vinyals, O.; Agapiou, J.;
  and kavukcuoglu, k. 2016.
\newblock Strategic Attentive Writer for Learning Macro-Actions.
\newblock In \emph{Advances in Neural Information Processing Systems},
  volume~29. Curran Associates, Inc.

\bibitem[{Vezhnevets et~al.(2017)Vezhnevets, Osindero, Schaul, Heess,
  Jaderberg, Silver, and Kavukcuoglu}]{feudalRL}
Vezhnevets, A.~S.; Osindero, S.; Schaul, T.; Heess, N.; Jaderberg, M.; Silver,
  D.; and Kavukcuoglu, K. 2017.
\newblock {F}e{U}dal Networks for Hierarchical Reinforcement Learning.
\newblock In \emph{Proceedings of the 34th International Conference on Machine
  Learning}, volume~70 of \emph{Proceedings of Machine Learning Research},
  3540--3549. PMLR.

\bibitem[{Zhang et~al.(2019)Zhang, Lipton, Pineda, Azizzadenesheli, Anandkumar,
  Itti, Pineau, and Furlanello}]{casual_state_rep}
Zhang, A.; Lipton, Z.~C.; Pineda, L.; Azizzadenesheli, K.; Anandkumar, A.;
  Itti, L.; Pineau, J.; and Furlanello, T. 2019.
\newblock Learning Causal State Representations of Partially Observable
  Environments.
\newblock \emph{CoRR}, abs/1906.10437.

\end{thebibliography}

\newpage
\clearpage
\section{Algorithm}

\begin{algorithm}
	\caption{CRADOL with Entropy Maximization}
	\label{alg:cradol}
	\begin{algorithmic}[1]
		 \State \text{Initialize inter-Q network parameters} $\psi_1, \psi_2$ 
		 \State \text{Initialize intra-Q network parameters} $\phi_1, \phi_2$
		 \State \text{Initialize intra-policy parameter $\theta$}
		 \State \text{Initialize beta-policy parameter $\curledv$}
		 \State \text{Initialize learning rates  $\lambda_{\psi}, \lambda_{\phi}, \lambda_{\theta}, \lambda_{\curledv}$}
		 \State \text{Initialize $\tau$ for soft target update}
         \State $D = []$ \Comment{Replay buffer initialization}
         \State Initialize internal states $h_{\psi_1}$, $h_{\psi_2}$, 
         $h_{\phi_1}$, $h_{\phi_2}$, 
         $h^{\text{RNN}}_{\omega}$
      \For{each iteration}
        \State Get initial observation $x$
        \State Get initial $\omega$ according to $\pi_\Omega$
        \For{each environment step}
            \State Get action $a$ and updated $h_{\omega}^{\text{RNN}}$ from $\pi_{\omega}$
            \State Get updated $h_{\psi_1}, h_{\psi_2}$, $h_{\phi_1}, h_{\phi_2}$ from $Q_{\Omega}, Q_U$
            \State Take action $a$ and observe $x'$ and $r$
            \State $D \gets D \cup \{x, \omega, a, r, x', h_{\psi_1}, h_{\psi_2}$ 
            \State $h_{\phi_1}, h_{\phi_2}, h_{\omega}^{\text{RNN}}\}$
            \If{$\omega$ \text{terminates in} $x'$ determined by $\beta_\omega$} 
                \State Get new $\omega$ according to $\pi_\Omega$ 
            \EndIf
      \EndFor
      \For{each gradient step} 
        \State{$\psi_i \gets \psi_i - \lambda_{\psi}\nabla_{\psi_i}J_{\Omega}(\psi_i)$ \text{for} $i=1,2$} 
        \State{$\Bar{\psi_i} \gets  \tau\psi_i +  (1-\tau)\Bar{\psi_i}$ \text{for} $i=1,2$} 
        \State{$\theta \gets \theta - \lambda_{\theta}\nabla_{\theta} J_{\omega}(\theta)$} 
        \State{$\phi_i \gets \phi_i - \lambda_{\phi}\nabla_{\phi_i} J_{U}(\phi_i)$ \text{for} $i=1,2$} 
        \State{$\Bar{\phi_i} \gets \tau\phi_i  +  (1-\tau)\Bar{\phi_i}$ \text{for} $i=1,2$} 
        \State{$\curledv \gets \curledv - \lambda_{\curledv}\nabla_{\curledv}J_{\beta}(\curledv)$} 
      \EndFor
    \EndFor
	\end{algorithmic}  
\end{algorithm}

\section{Option Learning Gradients}
Given a set of Markov options with stochastic intra-option policies differentiable in their parameters $\theta$, we denote the gradient $\nabla_{\theta}J_{\omega}(\theta)$ of the expected discounted return with respect to $\theta$ and initial condition $s_0, \omega_0$:
\begin{equation}
\smallsum_{s,\omega}\mu_{\Omega}(s,\omega|s_0, \omega_0)\!\smallsum_{a}\frac{\partial \pi_{\omega}(a|b_{\omega})}{\partial \theta}Q_{U}(b_U, \omega,a),
\end{equation} 
where $\mu_{\Omega}$ denotes the discounted weighting of $(s,\omega)$ along trajectories originating from $(s_0,\omega_0)$, with $b_\omega$ derived as denoted in \cref{fig:approach} and $x \sim O(s)$.
We expand the option-value function upon arrival with state abstraction as:
\begin{align}
\begin{split}
Q_{U}(b_{U},\omega,a)=r+\gamma\smallsum_{s'}P(s'|s,a)U(b_{\omega}', \omega),
\end{split}
\end{align}
where $U(b_{\omega}', \omega)$ equals to:
\begin{align}
\begin{split}
(1-\beta_{\omega}(b_{\omega}'))Q_{\Omega}(b_{\Omega}',\omega)\!+\! \beta_{\omega}(b_{\omega}')V_{\Omega}(b_{\Omega}').
\end{split}
\end{align}
Regarding the termination function, the gradient $\nabla_{\curledv}J_{\beta}(\curledv)$ of the expected discounted return objective with respect to $\curledv$ and the initial condition $(s_1, \omega_0)$ is:
\begin{equation}
\smallsum_{s',\omega}\!\mu_{\Omega}(s',\omega|s_1, \omega_0)\frac{\partial \beta_{\omega}(b_{\omega}')}{\partial \curledv}A_{\Omega}(b_{\Omega}', \omega),
\end{equation}
where $\mu_{\Omega}$ denotes the discounted weighting of $(s',\omega)$ along trajectories originating from $(s_1,\omega_0)$, and $A_{\Omega}$ is the advantage function over options. 

\section{Algorithm Clarifications}

\paragraph{Factored MDP assumption.}
We note that assuming the world as a factored MDP follows from a general Bayesian view of the world and enables representing a large, complex state space compactly. As noted by \citet{factoredmdps}, many real-world problems possess the context-specific structure and causal dependencies between states that can be exploited. For example, \citet{RIMs} leverages the structure of the world based on RIMs, resulting in impressive performance in many Atari environments. Thus, the factored MDP assumption does not limit the applicability of our method, and we expect that CRADOL also has advantages in Atari games.

\paragraph{Representation size.}
We clarify how CRADOL is consistent with our mathematical analysis of reducing the learning problem to be smaller than that of option critic. CRADOL performs abstraction on the representation level and not at the input level. 
We compare CRADOL to OC in terms of size of representation space.
For instance, let Figure 1 has $|\mathcal{A}|=4$, $|\mathcal{B}|=16$, and $|\Omega|=3$. Then, OC has a problem size of $|\Omega|(|A|)^{|\mathcal{B}|}=1.29\times 10^{10}$. But with both temporal and context-specific abstraction through CRADOL, the problem size is reduced to $\sum_{\omega\in\Omega} |A|^{|B_{\omega}|} = 4.80\times 10^3$.

\section{Additional Domain Details}
\paragraph{MiniGrid.} The observation is a partially observable view of the gridworld environment using a compact and efficient encoding, with 3 input values per visible grid cell and including 7x7x3 values in total. The agent is randomly initialized at each episode. For the Empty Room domain, an agent must go to the green goal location. For the MultiRoom domain, the agent must go to the green door to get to the green goal. Lastly, we test on the Key Domain in MiniGrid as described in the motivation. For the latter two domains, at the start of every episode, the structure of the grid also changes. Code for this domain can be found here: 
\textcolor{blue}{\url{{https://github.com/maximecb/gym-minigrid}}}.

\paragraph{Moving Bandit.} We modify this domain's termination condition to simulate sparse reward settings. Specifically, we terminate when the agent has reached the goal location, receiving a reward of 1. Code for this domain can be found at:  \textcolor{blue}{\url{https://github.com/maximilianigl/rl-msol}}.

\paragraph{Reacher.} To consider sparse reward settings, we make a minor change to the reward signal in this domain. A tolerance is specified to create a terminal state when the end effector approximately reaches the target. The tolerance creates a circular area centered at the target, which we specify as 0.003. We use the code from OpenAI Gym.

\newpage
\section{Additional Experiment Details}
We use the PyTorch library and GeForce RTX 3090 Graphics Card for running experiments. For specific versions of software, see the source code. 
We report hyperparameter values used across all methods, as well as episodic reward performance plots.

\paragraph{Note on hyperparameter $k$ choice.} 
In our experiments, we observed a tendency that increasing $k$ beyond a certain value (e.g., $k=10$) does not add significant benefit to reducing the complexity of the learning problem and resulting final evaluation reward. 
If $k$ was set to be too small, CRADOL would not perform well because only a small amount of factorization is being applied.

\begin{table}[b!]
\begin{center}
      \centering
      \caption{MiniGrid EmptyRoom}
      \begin{tabular}{l|r}
        \toprule Hyperparameter & Value \\
        \midrule Batch Size  & 100 \\
                Learning Rate $\lambda$ &  0.0003 \\
                Entropy Weight &  0.005 \\
                Options $|\Omega|$ & 3 \\
                Mechanisms $K$ & 4 \\
                Top-k $k$ & 3 \\
                Hidden Size per RNN & 6 \\
                Value Size $v$ & 32 \\
                Discount Factor $\gamma$ & 0.95  \\
        \bottomrule
      \end{tabular}
\end{center}
\begin{center}
      \centering
      \caption{MiniGrid Multi-Room \& DoorKey}
      \begin{tabular}{l|r}
        \toprule Hyperparameter & Value \\
        \midrule Batch Size  & 100 \\
                Learning Rate $\lambda$ &  0.0005 \\
                Entropy Weight &  0.001 \\
                Options $|\Omega|$ & 3 \\
                Mechanisms $K$ & 4 \\
                Top-k $k$ & 3 \\
                Hidden Size per RNN & 6 \\
                Value Size $v$ & 32 \\
                Discount Factor $\gamma$ & 0.95  \\
        \bottomrule
      \end{tabular}
    \end{center}
\begin{center}
      \centering
      \caption{Moving Bandit}
      \begin{tabular}{l|r}
        \toprule Hyperparameter & Value \\
        \midrule Batch Size  & 100 \\
                Learning Rate $\lambda$ &  0.005 \\
                Entropy Weight &  0.005 \\
                Options $|\Omega|$ & 3 \\
                Mechanisms $K$  & 4 \\
                Top-k $k$ & 3 \\
                Hidden Size per RNN & 6 \\
                Value Size $v$ & 16 \\
                Discount Factor $\gamma$ & 0.95  \\
        \bottomrule
      \end{tabular}
    \end{center}
\begin{center}
      \centering
      \caption{Reacher}
      \begin{tabular}{l|r}
        \toprule Hyperparameter & Value \\
        \midrule Batch Size  & 100 \\
                Learning Rate $\lambda$ &  0.001 \\
                Entropy Weight &  0.001 \\
                Options $|\Omega|$ & 3 \\
                Mechanisms $K$ & 6 \\
                Top-k $k$ & 4 \\
                Hidden Size per RNN & 6 \\
                Value Size $v$ & 64 \\
                Discount Factor  $\gamma$ & 0.95  \\
        \bottomrule
      \end{tabular}
\end{center}
  \end{table}

\begin{figure}[b!]
  \begin{center}
  \centerline{\includegraphics[width=0.6\linewidth]{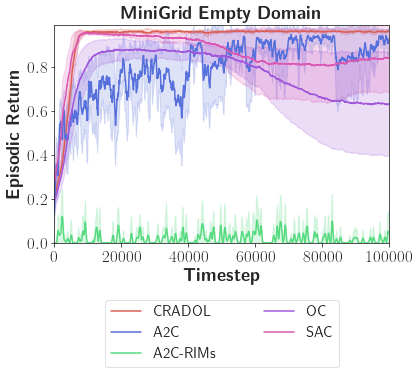}}
  \caption{MiniGrid-Empty-Random-6x6-v0}
  \label{empty_random_graph}
  \end{center}
  \begin{center}
  \centerline{\includegraphics[width=0.6\linewidth]{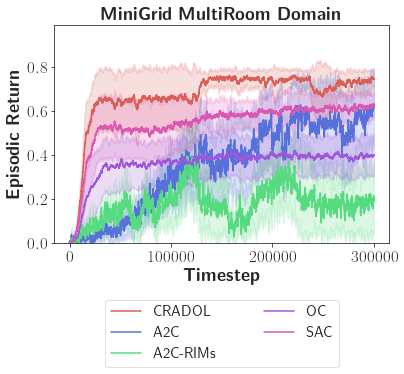}}
  \caption{MiniGrid-MultiRoom-N2-S4-v0}
  \label{multi_room_graph}
  \end{center}
  \begin{center}
  \centerline{\includegraphics[width=0.6\linewidth]{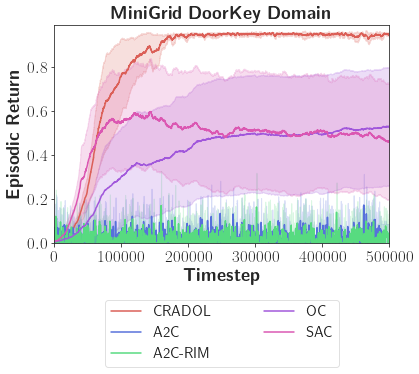}}
  \caption{MiniGrid-DoorKey-6x6-v0}
  \label{door_key_graph}
  \end{center}
  \begin{center}
  \centerline{\includegraphics[width=0.6\linewidth]{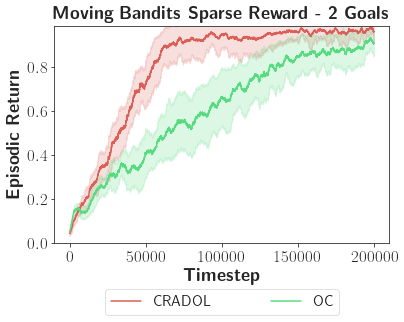}}\caption{Moving Bandits with 2 goal locations.}
  \label{auc_bandit_2_goals}
  \end{center}
\end{figure}

\end{document}